\documentclass[runningheads]{llncs}
\usepackage{graphicx}
\usepackage{xcolor}
\usepackage{float}
\usepackage{verbatim}

\begin{document}
\title{Quantum Transfer Learning for Acceptability Judgements}
\author{Giuseppe Buonaiuto\inst{1} \and
Raffaele Guarasci\inst{1,*}
\and
Aniello Minutolo\inst{1} \and
Giuseppe De Pietro\inst{1}\and
Massimo Esposito\inst{1}}
\authorrunning{F. Author et al.}
\institute{National Research Council of Italy (CNR), Institute for High Performance Computing and Networking (ICAR), Naples, 80131, Italy \\
\inst{*}\email{raffaele.guarasci@icar.cnr.it}}
\maketitle       % typeset the header of the contribution
\begin{abstract}
Hybrid quantum-classical classifiers promise to positively impact critical aspects of natural language processing tasks, particularly classification-related ones. Among the possibilities currently investigated, quantum transfer learning, i.e., using a quantum circuit for fine-tuning pre-trained classical models for a specific task, is attracting significant attention as a potential platform for proving quantum advantage. 

This work shows potential advantages, both in terms of performance and expressiveness, of quantum transfer learning algorithms trained on embedding vectors extracted from a large language model to perform classification on a classical Linguistics task: acceptability judgments.
Acceptability judgment is the ability to determine whether a sentence is considered natural and well-formed by a native speaker. 
The approach has been tested on sentences extracted from ItaCoLa, a corpus that collects Italian sentences labeled with their acceptability judgment. The evaluation phase shows results for the quantum transfer learning pipeline comparable to state-of-the-art classical transfer learning algorithms, proving current quantum computers' capabilities to tackle NLP tasks for ready-to-use applications. Furthermore, a qualitative linguistic analysis, aided by explainable AI methods, reveals the capabilities of quantum transfer learning algorithms to correctly classify complex and more structured sentences, compared to their classical counterpart. This finding sets the ground for a quantifiable quantum advantage in NLP in the near future. 

\end{abstract}
\section{Introduction}

Deep learning models have fostered significant advances in Natural Language Processing (NLP), especially in machine translation, text classification, and syntactic analysis. These improvements are primarily due to Transformers models, such as BERT, which changed the previously adopted paradigm thanks to the introduction of context-aware mechanisms. However, training large and complex language models requires massive amounts of data and resources. Hence, when dealing with specific NLP tasks, it is customary to use pre-trained models and use them carefully adapted for solving the task. The adaptation is generally given by a new deep learning algorithm to be trained that uses the outputs of the pre-trained model as input features. Taking a model trained to do one task and then fine-tuning it to work on a related but different task is at the essence of what is called Transfer Learning. Transfer learning has been widely used in NLP to improve various tasks \cite{ruder2019transfer}, ranging from cross-lingual approaches to overcome limitations of low-resource languages\cite{schuster2018cross,guarasci2021assessing} or trying to infer linguistic knowledge between typologically different languages \cite{guarasci2022bert,kim2017cross}; domain adaptation \cite{ma2019domain} in machine-translation \cite{shah2018adversarial} or named-entity recognition related tasks \cite{ruder2017knowledge}.

%Recently, there were significant advancements in all Natural Language Processing (NLP) tasks~\cite{vaswani2017attention,devlin2018BERT,radford2019language,clark2020electra}, ranging from machine translation~\cite{zhu2020incorporating}, text classification~\cite{sun2019fine}, coreference resolution~\cite{lee2017end,guarasci2021electra} or multi-language syntactic analysis~\cite{guarasci2022BERT,guarasci2021assessing,chi2020finding}. This is primarily due to the explosion of neural language models based on deep learning architecture, especially Transformers-based models such as BERT \cite{devlin-etal-2019-BERT}. However, this improvement comes with an increasing complexity of models, which requires a considerable amount of data and computation resources to be efficiently trained \cite{brown2020language,floridi2020gpt}. Besides, there are open issues related to what these models learn about language, how they encode this information \cite{Jawahar2019}, and how much of it is interpretable \cite{jiang2020can}. 

Quantum machine learning (QML) is gaining attention as an alternative approach that exploits powerful aspects borrowed from quantum mechanics to overcome the computational limitations of current approaches. A strand derived from QML is the sub-field of Quantum NLP (QNLP) \cite{coecke2010mathematical}, which aims to solve natural language-related tasks using quantum properties or applying algorithms derived from quantum theory or testing new approaches using real quantum hardware.

However, quantum-based approaches currently suffer the limitations of available hardware, and there are open issues about scalability because the quantum circuit can be considered a linear or, at most, sub-linear model in the feature space \cite{schuld2021quantum}. There is an ongoing research effort concerning the encoding of classical data into quantum computers, which in principle allows to extract information and make computations more efficiently than classical computers, but with a strong backside when considering large datasets or data represented as large vectors in feature space. Nowadays, hardware has a limited amount of qubits, and when the number of qubits is sufficient for embedding data, the lack of fault tolerance, i.e., the quantum noise associated with the computation, hinders the performances and destroys the potential advantages of quantum computing. To overcome these limitations, novel hybrid approaches combining classic pre-trained models and quantum techniques have been proposed \cite{li2022quantum,li2023adapting}. This type of approach offers the advantage of implementing specific layers of models on a quantum device, while classical models for non-linear operations handle intermediate results. A successful example of such hybrid approaches is the so-called classical–quantum transfer learning \cite{mari2020transfer}. It encodes input features in a multi-qubit state, and then a quantum circuit transforms and measures such features. In this pipeline, output probabilities are projected to the task label space, and losses are backpropagated to update parameters.

Starting from the approach proposed in \cite{li2023adapting}, this work proposes a hybrid transfer learning model for QNLP applied to a binary classification task. It uses two pre-trained models, namely BERT and ELECTRA, and fine-tunes them to perform the classification. This approach has been tested on current noisy intermediate-scale quantum (NISQ) machines \cite{torlai2020machine}. It exploits the advantage and robustness of classic pre-trained language features already well-known in the literature \cite{qiu2020pre} and integrates quantum encodings to turn classical data into quantum state and parametrized variational circuits \cite{Li2021} to perform classification. 
In this sense, the quantum model used for the classification task falls into the broad category of variational quantum classifiers (VQCs) \cite{chen2020hybrid}. Here, the VQCs algorithms are designed to perform binary classification on acceptabiliy judgments, a fundamental task in Theoretical and Computational Linguistics \cite{sprouse2013assessing,lau-etal-2015-unsupervised,linzen2016assessing} and computation, which has gained much popularity in recent years in the field of NLP \cite{warstadt-etal-2019-neural,Linzen2019WhatCL}. This particular task has been chosen because QNLP has proven particularly effective in binary classification \cite{correia2021grover,meichanetzidis2020grammar,sordoni2013modeling,qnlp_numta}. 

In NLP, acceptability judgments refer to assessing whether a given sentence is grammatically correct, semantically meaningful, or perceived as natural by native speakers. Many approaches in recent years have addressed this task, running into numerous critical issues \cite{linzen2016assessing,sprouse2013empirical}. Foremost is the scarcity of available resources. Obtaining annotated data on acceptability judgments is an onerous and time-consuming activity that requires expert native speakers \cite{sprouse2013comparison}. This problem becomes even more acute when dealing with low-resource languages other than English.

For this work, which focuses on the Italian language, the dataset chosen has been ItaCoLa\cite{trotta-etal-2021-monolingual-cross}. It is the largest existing resource for this task in Italian, collecting sentences labeled with their judgments by expert linguists.
Here, it is demonstrated that the quantum transfer learning approach for these topics can give enhanced accuracy in classification without being time and resource-consuming. The quantum advantage given by the quantum classifier is inferred by direct comparison with the state-of-the-art classical deep learning models used for the ItaCola Acceptability Judgements (namely LSTMs and ITABert). 

The paper is organized as follows: in section \ref{sec:related}, the research works available in the literature related to what is presented are described. In section \ref{sec:methods} the dataset and applied methodologies are presented. Section \ref{sec:Experimental phase}  provides details about the experimental phase, such as the parameter in use and details about the computational pipelines. Then, in section \ref{sec:results}, the results obtained are exposed, relevant aspects are discussed, and finally, overall conclusions are drawn in section \ref{sec:conclusion}.

\section{Related Work}
\label{sec:related}

\subsection{Quantum Machine Learning}

Numerous examples have received increasing interest in recent years concerning the adaptation of classical machine learning algorithms through the use of properties and techniques borrowed from quantum mechanics. One notable approach is Quantum Support Vector Machines (QSVMs), which aim to enhance the performance of traditional Support Vector Machines (SVMs) by utilizing quantum algorithms. QSVMs have shown promising results in various text classification tasks, including sentiment analysis, topic classification, and document classification. Another area of research is quantum-inspired algorithms for text classification. Quantum-inspired algorithms, such as Quantum-Inspired Genetic Algorithm (QGA) and Quantum-Inspired Particle Swarm Optimization (QPSO), draw inspiration from quantum mechanics and apply quantum-like principles to improve traditional optimization techniques. These algorithms have been explored in the context of feature selection and parameter optimization for text classification, demonstrating their potential to enhance classification accuracy and efficiency.
Moreover, quantum embeddings have gained attention as a means to represent and analyze textual data. Quantum embeddings leverage the concepts of quantum superposition and entanglement to capture semantic relationships between words or documents. These embeddings aim to capture more nuanced and context-dependent information than traditional word embeddings. By utilizing quantum representations, classification models can benefit from enhanced semantic understanding, improving performance in various NLP tasks.

Furthermore, quantum machine learning algorithms, such as Quantum Neural Networks (QNNs) and Quantum Boltzmann Machines (QBMs), have been explored in text classification. These quantum-inspired models leverage the unique computational capabilities of quantum systems to perform complex computations efficiently. Although still in its early stages, quantum machine learning holds the potential to address the computational challenges associated with large-scale language data and to provide more robust models for classification tasks.

Among the most immediate and promising applications of quantum machine learning algorithm, classical-to-quantum (CQ) \cite{Mari2} transfer learning is attracting growing attention as it promises to show quantum advantage on specific tasks, using both the representation power of state-of-the-art pre-trained deep learning models and the expressivity of quantum circuit, which are constructed to tackle the problems to be solved. 
%Among the most interesting application of classical to quantum transfer learning it is worth to notice 
\subsection{Quantum Transfer Learning}
Typically, transfer learning indicates a set of techniques in Artificial Intelligence where knowledge acquired from a specific task is transferred to solve a different problem. This idea has been successfully applied in various contexts, from image classification to sentiment analysis. Generally, a pre-trained network is used, trained on a particular dataset, from which the last layer is removed. The truncated pre-trained network is then interpreted as a feature extractor on the new dataset. The feature vectors are then used to train a dedicated neural network designed to address the task. Quantum transfer learning can be realized within two frameworks: quantum-transfer learning algorithms use feature vectors extracted from a trained quantum machine learning algorithm and then fed into a quantum neural network. Instead, a classical-quantum transfer learning algorithm uses the features extracted from a classical neural network, which are then encoded into a task-tailored quantum neural network. In this work, the latter approach is used, as it is convenient for elaborating the highly informative feature vectors extracted from large language models (BERT and ELECTRA) and then using a quantum classifier to infer the relation between vectors and the target classes. 

\subsection{Quantum Natural Language Processing}

The growing interest in exploring the potential of QML in language-related tasks has led to the birth of the QNLP. As a sub-field of QML, QNLP offers a new paradigm focused on processing and analyzing language data by leveraging the principles of quantum mechanics.

At the dawn of the QNLP, the first approaches proposed were only at the theoretical level. They proposed algorithms based on quantum theory that are potentially implementable on quantum hardware but not tested on actual data \cite{zeng2016quantum,coecke2020foundations}. These approaches covered different tasks of NLP, from generic aspects of sentence representations using distributional or compositional properties to specific tasks \cite{correia2021grover,abbaszade2021application}. 

More in theme with the purpose of this paper, however, are the approaches that were tested on real datasets using classical hardware (quantum-inspired approaches) or currently available quantum machines (quantum-computer approaches).
Quantum-inspired approaches are typically structured in a specific manner. They start with the classic model and then integrate the advancements of quantum mechanics to enhance their performance, proving that a quantum approach can surpass the current state-of-the-art. Some of these studies have been evaluated against benchmark datasets found in the literature, while others have utilized custom-made samples for their assessments.

Finally, quantum-computer approaches proposed alternatives to simple NLP tasks that can be performed on quantum hardware (NISQ). Currently, experiments are limited to small-to-medium scale datasets due to the current limits of quantum hardware. Notice that several studies have investigated the application of quantum computing in classification tasks within NLP; a detailed review is described in \cite{guarasci2022quantum}.

An alternative way proposed to push through the limitations inherent in the scalability of such experiments is represented by hybrid approaches.
A hybrid classical-quantum scheme using a quantum self-attention neural network (QSANN) has been recently proposed\cite{li2022quantum}. Even if it introduces the possibility of non-linearity, significantly improving over other QNLP models\cite{lloyd2020quantum}, this approach is limited by the continuous switching between quantum and classical hardware at each self-attention layer needed to run the network.

\cite{li2023adapting} proposes a more viable approach to solve the low non-linearity issue for QNLP models using the classical-quantum transfer learning paradigm \cite{mari2020transfer}. Using the classical-quantum transfer mechanism and pre-trained quantum encodings seems the most promising approach to develop scalable QNLP models, paving the way for the possibility of being implemented on real quantum hardware. Given this high potential and the increasing appeal of this approach, it was chosen for use in this work.

\subsection{Acceptability Judgements Task}

Understanding and accurately predicting acceptability judgments is crucial for various NLP applications, including grammar correction, machine translation, and automated dialogue systems. Therefore, there are some open issues. Firstly, acceptability is often subjective and context-dependent, varying across different languages. Secondly, various linguistic phenomena affect such judgments, such as syntax, semantics, pragmatics, and discourse. Therefore, models that have addressed this task so far have had to cope with these factors, capturing fine-grained linguistic features \cite{linzen2018reliability} and generalizing across diverse linguistic contexts using cross-lingual approaches \cite{cherniavskii2022acceptability}. Furthermore, labeled data for training such models is typically scarce and costly to obtain, as it requires expert annotation or crowd-sourcing efforts. This scarcity of labeled data necessitates using transfer learning and other techniques to leverage pre-trained language models and enhance the performance of acceptability judgment prediction models. 

% CAPPELLO
In recent years, significant progress has been made in developing neural network architectures, such as recurrent neural networks (RNNs), convolutional neural networks (CNNs), and transformer models, which have shown promising results in predicting and classifying acceptability judgments. These models leverage sentences' syntactic structure and semantic content, capturing the contextual information necessary for accurate acceptability predictions.

%COLA 
Automatically assessing acceptability tasks gained great popularity since the release of the CoLa corpus \cite{warstadt-etal-2019-neural}, the first large-scale corpus of English acceptability, containing more than 10k sentences from linguistic literature. 

The CoLA corpus has been presented with several experiments to assess the performance of neural networks on a novel binary acceptability task. Furthermore, it has been included in the GLUE dataset \cite{wang-etal-2018-glue}, a very popular multi-task benchmark for English natural language understanding, and an acceptability challenge has been launched on Kaggle,\footnote{\url{https://www.kaggle.com/c/cola-in-domain-open-evaluation/}}. For such reasons, the number of studies dealing with binary acceptability has remarkably increased. However, proposed approaches have often used different metrics for evaluation, so comparisons cannot always be made.

%Unfortunately, most studies using GLUE report accuracy instead of MCC, making it difficult to identify the best approach. Nevertheless, all top-ranked systems rely on variations of transformer-based models, including ALBERT \cite{DBLP:conf/iclr/LanCGGSS20} (69.1 Accuracy) and StructBERT \cite{DBLP:conf/iclr/0225BYWXBPS20} (69.2 Acc.). More recently, also reformulating acceptability as an entailment task and using smaller language models to few-shot fine-tuning has shown a great potential \cite{wang2021entailment}, outperforming existing BERT-based approaches (86.4 Acc.). 

%The best performance was achieved with a pooling classifier and ELMo-style embeddings (0.341 MCC). Matthews Correlation Coefficient (MCC) was chosen as an evaluation measure because it is more appropriate than F1 or accuracy for binary classification with unbalanced data \cite{MATTHEWS1975442}. More recently, \cite{DBLP:journals/corr/abs-1901-03438} has compared a BiLSTM baseline with the performance achieved by transformer encoders such as GPT and BERT. The best approach is obtained by fine-tuning BERT$_{large}$ (MCC = 0.582).
%Other approaches, instead, focus on unsupervised learning, for example \cite{lau-etal-2015-unsupervised,lau-etal-2020-furiously} compare different types of language models to infer the probability of a sentence, which is then mapped onto acceptability. 

%% ITACOLA &  FRIENDS

Starting with the same methodology introduced in COLA, similar resources have been released in recent years in various typologically different languages, ranging from Italian \cite{trotta-etal-2021-monolingual-cross,bonetti2022work} - which is the subject of this work - Norwegian \cite{jentoft2023nocola}, Swedish \cite{volodina-etal-2021-dalaj} and Russian \cite{mikhailov2022rucola}, Japanese \cite{someya2023jblimp}, and Chinese\cite{xiang-etal-2021-climp}.

% ALTRI DATASETS ESISTENTI E INUTILI
Notice that other acceptability datasets already existed in the literature, but these were small resources that arose for purely theoretical purposes or within psycholinguistic experiments \cite{sprouse2013empirical,lau2014measuring,marvin2019targeted}. Regarding languages other than English, \cite{linzen2018reliability} evaluate informal acceptability judgments on Hebrew and Japanese. A similar study has been conducted in French \cite{feldhausen2020testing} and in Chinese \cite{chen2020assessing}. A small dataset in the context of Evalita 2020 evaluation campaign on complexity and acceptability fort the Italian language (AcComplIt task) \cite{DBLP:conf/evalita/BrunatoCDMVZ20} has been released.

\section{Materials and Methods}
\label{sec:methods}

\subsection{Neural Language Models}

For this work, two Neural Language Models (NLMs) have been considered, namely BERT and ELECTRA.

\subsubsection{BERT}
% cappello bert
Among NLMs in literature, BERT\cite{Devlin2019} is the most widely used due to its efficiency and high performance. 
Generally speaking, BERT is a multi-layer bidirectional architecture based on the original Transformer encoder \cite{vaswani2017attention}, pre-trained on large-scale unlabeled text via two training goals, i.e., masked language modeling and next sentence prediction. 

A pre-trained BERT model typically provides a powerful context-dependent sentence representation that can be successively adapted to a downstream NLP task through a fine-tuning procedure according to different needs.
The fine-tuning procedure requires configuring several hyper parameters whose values directly influence the results that can be obtained. 

The $BERT_{base}$ model consists of $12$ hidden layers, each one having $768$-hidden dimensional states and $12$ attention heads, with a total of $110$M parameters. The $BERT_{base}$ model accepts input sequences of words, with a maximum length of $512$. Each layer of the model encodes a distinct embedded representation of the input words, which can be leveraged for various NLP tasks, including the syntactic probe discussed in this paper.

Masked language modeling involves randomly masking a percentage of words in the training corpus. By doing so, the pre-trained model learns to encode information from both directions of the sentences and simultaneously predict the masked words. The input vocabulary can be either \textit{cased} or \textit{uncased}, resulting in two different pre-trained models. The flexibility offered by bidirectional analysis simultaneously allows, on the one hand, to maintain a large generating capacity through the inner layers of the deep constituent network and, on the other hand, to use the outer layers of the network to adapt to the specific task through the fine-tuning phase, is what has allowed BERT to be the benchmark model in the literature in recent years.

BERT expects that each input sequence of words starts with a unique token \textit{[CLS]}, used to obtain in output a vector of size $H$, i.e., the size of the hidden layers, representing the entire input sequence. Moreover, the unique token \textit{[SEP]} must be placed within the input sequence at the end of each sentence. 

Given an input sequence of words $t = (t_1, t_2, …, t_m)$, the output of BERT is $h = (h_0, h_1, h_2, …, h_m)$ where $h_0 \in R^H$ is the final hidden state of the special token \textit{[CLS]} and provides a pooled representation for the full input sequence, while $h_1, h_2, …, h_m$ are the final hidden states of other input tokens.

To fine-tune BERT for classifying input sequences of words into $K$ different text categories, the final hidden state $h_0$ can be used to feed a classification layer, with a subsequent softmax operation to turn the scores of each text category into likelihoods \cite{sun2019fine}:

\begin{equation}
    P = softmax(CW^T)
\end{equation}

where $W \in R^{KxH}$ is the parameter matrix of the classification layer.

\subsubsection{ELECTRA}
The second NLM taken into account is ELECTRA \cite{clark2020electra}, since it has shown a better ability to capture contextual word representations outperforming, in its downstream performance, other models, like BERT, given the same model size, data, and compute \cite{rogers2020primer}.

Generally speaking, ELECTRA is a pre-training approach that trains two transformer models, namely the generator $G$ and the discriminator $D$. 
The role of the model $G$ is to replace tokens in a sequence and is, therefore, usually trained as a masked language model. The model $D$, which is typically the ELECTRA model of interest, tries instead to identify which tokens were replaced by $G$ in the sequence, and it may be a BERT-based model, virtually any model producing an output distribution over tokens.

In particular, for a given input sequence, where some tokens are randomly replaced with a special \textit{[MASK]} token, $G$ is trained to predict the original tokens for all masked ones, after which $G$ generates a fake input sequence for $D$ by replacing the [MASK] tokens with fakes. Finally, $D$ is given the fake sequence as input and is trained to predict whether their tokens are original or \textit{fake}. This approach, replaced by token detection (RTD), allows the use of a minor number of examples without losing performance.

\begin{figure}[ht]
\centering
\includegraphics[width=0.70\textwidth]{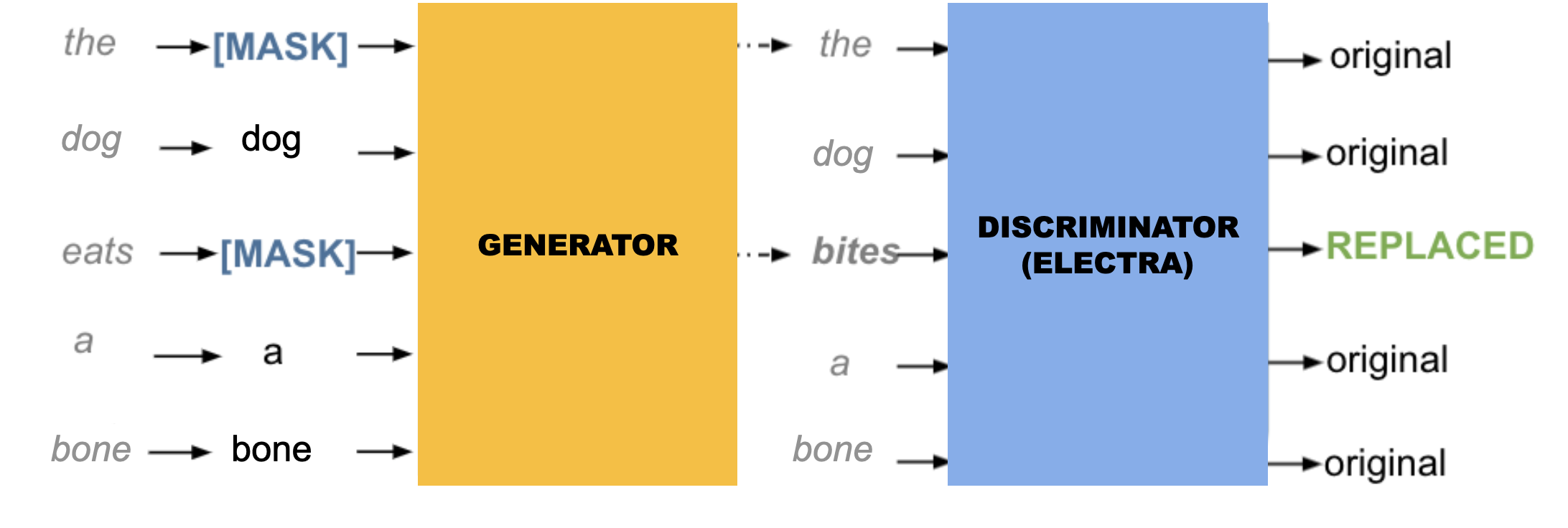}
\caption{ELECTRA overview with replaced token detection.}
\label{fig:electra}
\end{figure}

More formally, given an input sentence $s$ of raw text $\chi$, composed by a sequence of tokens $s=w_1, w_2, \dots, w_n$ where $w_t$ ($1\leq t\leq n$) represents the generic token, 
both $G$ and $D$ firstly encode $s$ into a sequence of contextualized vector representations $h(s) = h_1, h_2, \dots, h_n$. 

Then, for each position $t$ for which $w_t = [MASK]$, the generator $G$ predicts, through a softmax layer, the probability to generate a specific token $w_t$:

\begin{equation}
    p_G(w_t|s) = \frac{e(w_t)^T h_G(s)_t}{\sum_{w'}exp(e(w')^T h_G(s)_t)}
\end{equation}

where $e(\cdot): w_t \in s \rightarrow R^{dim}$ is the embedding function, and \textit{dim} the chosen embedding size. 

The discriminator $D$ predicts, via a sigmoid layer, if $w_t$ is original or "fake":

\begin{equation}
    D(s,t) = sigmoid(e(w_t)^Th_D(s)_t)
\end{equation}

During the pre-training, the following combined loss function is minimized:

\begin{equation}
    \min_{\theta_G,\theta_D}\sum_{s \in \chi} \mathcal{L}_{Gen}(s,\theta_G)+\lambda\mathcal{L}_{Dis}(s,\theta_D)
\end{equation}

where $\mathcal{L}_{Gen}$ and $\mathcal{L}_{Dis}$ are the loss functions of $G$ and $D$, respectively. 

At the end of the pre-training, $G$ is discarded and only $D$ is effectively used for fine-tuning on the specific task.

Masked language modeling pre-training methods such as BERT corrupt the input by replacing some tokens with [MASK] and then training a model to reconstruct the original tokens. While they produce good results when transferred to downstream NLP tasks, they generally require large amounts of computing to be effective. As an alternative, replaced token detection is a more sample-efficient pre-training task that corrupts the input by replacing some tokens with plausible alternatives sampled from a small generator network instead of masking the input. The main reason ELECTRA efficiency results improved concerning BERT-like NLMs is that predictions are calculated not only over masked tokens but also for the other tokens in the input sequence, and, thus, the discriminator loss can be calculated over all input tokens. It allows using a minor number of examples without losing in performance.

\subsubsection{Pre-trained models in the Italian language}

\begin{comment}
Notice that, training such a large neural architecture demands considerable time and computational resources, for this reason various pre-trained BERT models in different languages have been made available in the literature.   
\end{comment}

For the purpose of this work, since training an NLM demands considerable time and computational resources, two pre-trained versions of BERT and ELECTRA in the Italian language have been adopted.

As a pre-trained version of BERT, the cased and XXL version of the \textit{dbmdz} Italian BERT model\footnote{https://huggingface.co/dbmdz/bert-base-italian-xxl-cased} has been chosen. This model has been trained with cased vocabularies with an initial sequence length of 512 subwords for ~2-3M steps. The source data consists of an "XXL" corpus composed of a Wikipedia dump, various texts from the OPUS\footnote{http://opus.nlpl.eu/} corpora collection, and data from the Italian part of the OSCAR\footnote{https://traces1.inria.fr/oscar/} corpus. Thus, the final training corpus has a size of 81GB and 13,138,379,147 tokens.

As pre-trained version of ELECTRA, the cased and XXL version of the \textit{dbmdz} Italian ELECTRA model\footnote{https://huggingface.co/dbmdz/electra-base-italian-xxl-cased-discriminator} has been chosen. This model has been trained on the "XXL" corpus for 1M steps in total using a batch size of 128. The training procedure followed is the one used for BERTurk\footnote{https://github.com/stefan-it/turkish-bert/tree/master/electra}.

Notice that in recent years, other pre-trained versions of BERT and ELECTRA in the Italian language have been made available \cite{Polignano_2019}. However, they are usually trained on social media and Twitter corpora, since they present an often non-standard linguistic variety at the syntactic and lexical level. For this reason, the \textit{dbmdz} models, which have been trained on a corpus formed by documents written in standard language, are the most appropriate choice in a task that aims to assess acceptability.

\subsection{ItaCola Dataset}

The dataset used for this work is ItaCoLA, the Italian Corpus of Linguistic Acceptability \cite{trotta-etal-2021-monolingual-cross}. This corpus was designed to represent a broad range of linguistic phenomena while distinguishing between sentences that are considered acceptable and those that are not. The process used to create the corpus was modeled as closely as possible on the methodology employed for the English CoLA dataset \cite{warstadt-etal-2019-neural}. ItaCoLA consists of approximately 9,700 sentences drawn from various sources that cover numerous linguistic phenomena.

\begin{table}[h!]
\centering
\begin{tabular}{|l|l|}
\hline
Label & Sentence \\ \hline
0 & \begin{tabular}[c]{@{}l@{}}*Max è andato nella sua l'anno scorso casa. \\ (*Max went to his last year house)\end{tabular} \\ \hline
1 & \begin{tabular}[c]{@{}l@{}}l'ufficiale ha voglia di baciare Rosa \\ (The officer want to kiss Rose)\end{tabular} \\ \hline
0 & \begin{tabular}[c]{@{}l@{}}*Quella storia mi hanno spaventato. \\ (*That story have scared me)\end{tabular} \\ \hline
1 & \begin{tabular}[c]{@{}l@{}}Quella storia mi ha spaventato. \\ (That story has scared me)\end{tabular} \\ \hline
\end{tabular}
\caption{Example sentences from the ItaCoLA dataset. 1 = acceptable, 0 = not acceptable. The symbol $*$ is conventionally used in Linguistics to mark unacceptable sentences.} \label{tab:corpus}
\end{table}

The acceptability annotation for these sentences is based on Boolean judgments formulated by experts who authored the different data sources. It ensures robustness and simplifies classification. The sentences are sourced from various linguistic publications spanning four decades, transcribed manually, and released in digital format. The sources include theoretical linguistics textbooks and scientific articles on phenomena such as idiomatic expressions, locative constructions, and verb classification. An example of how the data are structured in the corpus is shown in Table \ref{tab:corpus}.

ItaCola is divided into training, validation, and test split, including 7,801, 946, and 975 examples. Notice that each split is balanced concerning sources composing the whole dataset, and the acceptability/not acceptability ratio is preserved. 

\begin{comment}
In \cite{trotta-etal-2021-monolingual-cross} a baseline using LSTM wit FasText embeddings was established. In addition, a classifier using an Italian version of BERT \cite{devlin-etal-2019-BERT} and fine-tuned using ItaCoLA training dataset was tested. Performances of these classifiers are shown in Table \ref{tab:baseline}. Evaluations were performed using two metrics: Accuracy, the most used measure used to evaluate acceptability on the GLUE benchmark and Matthews Correlation Coefficient (MCC) \cite{peters2018deep}, which is a correlation measure for Boolean variables mainly fitted when evaluating unbalanced binary classifiers. 

\begin{table}[]
\centering
\begin{tabular}{|l|c|c|}

\hline
\textbf{Model} & \textbf{Accuracy} & \textbf{MCC} \\ \hline
LSTM & $0.794$ & $0.278 \pm 0.029$ \\ \hline
BERT-Classic & $0.904$ & $0.603 \pm 0.022$ \\ \hline
\end{tabular}
\caption{Classification results on the ItaCoLA test set using LSTM and BERT.} \label{tab:baseline}
\end{table}
    
\end{comment}

\section{Experimental Phase}
\label{sec:Experimental phase}
\begin{figure*}[h!]
\centering
\includegraphics[width=\textwidth]{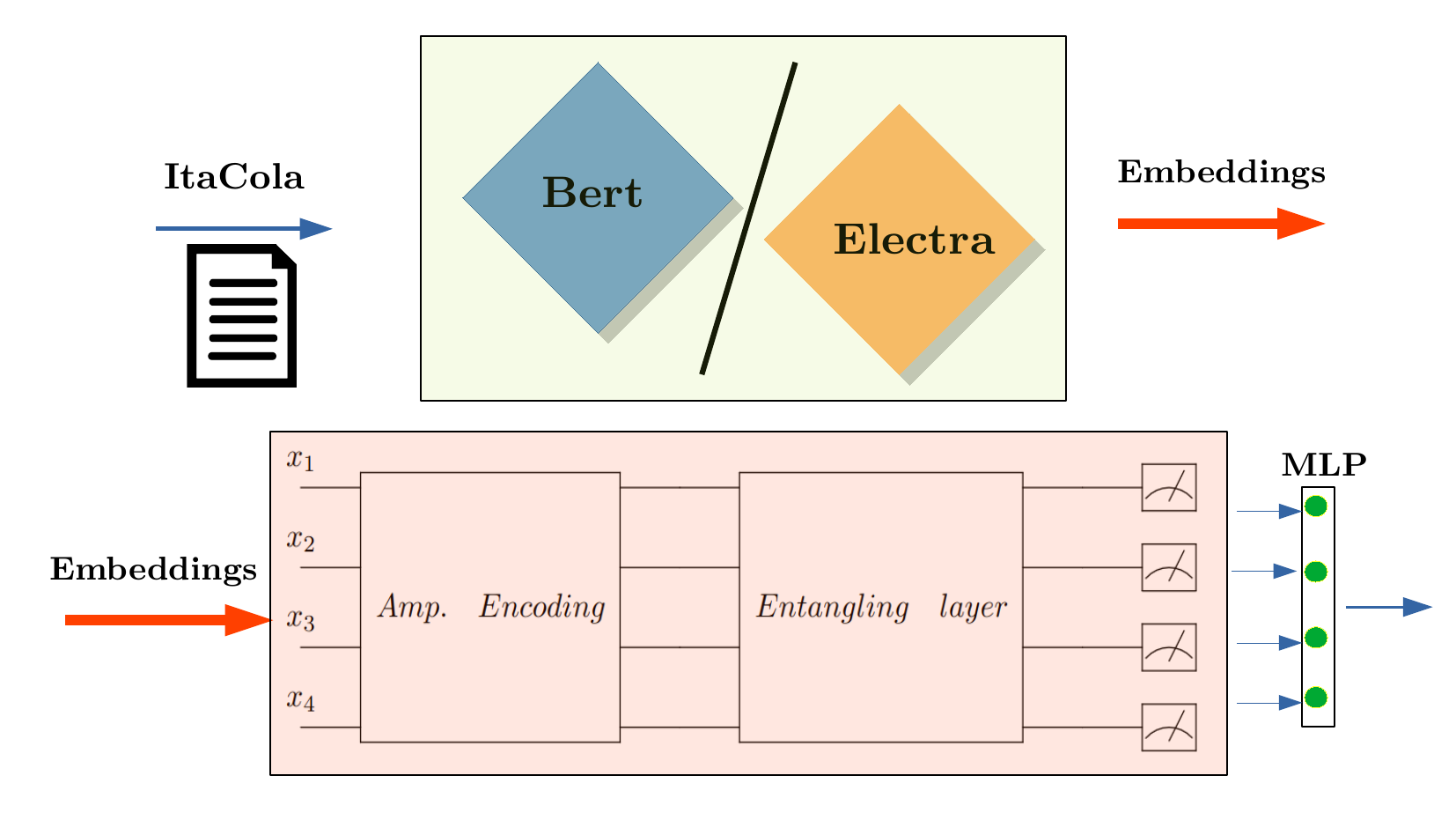}
\caption{\textbf{Schematic of the quantum transfer learning scheme adopted.} Sentences from ItaCola dataset are tokenized and then embeddings from BERT and ELECTRA (both pre-trained) are obtained for each data point. The embeddings are then encoded in the parametrized quantum circuit via amplitude encoding. The results of the measurement on the quantum states are then fed into a multi-layer perceptron (MLP) through which the classification is performed. }
\label{fig:scheme}
\end{figure*}

In this study, two quantum classification pipelines are employed. A schematic of each model is presented in Figure \ref{fig:scheme}. These quantum pipelines are compared with the corresponding classical equivalent. A detailed explanation is provided subsequently.
\subsection{Quantum pipelines}
This section outlines the methodology for incorporating Italian BERT and ELECTRA embeddings in the Quantum Natural Language Processing (QNLP) pipeline for the classification task of acceptability judgments. For convenience, The two transfer learning pipelines are called \textbf{BERT-Quant} and \textbf{ELECTRA-Quant} in the following sections.
The proposed pipeline is articulated in the following steps:
\begin{itemize}
    \item \textbf{Data Preprocessing}: Sentences extracted from ItaCola must be preprocessed and prepared for embedding generation.
    \item \textbf{Embedding Generation}: For this work, pre-trained BERT and ELECTRA  embeddings have been used. Feature vectors extracted via both models considered have $768$ real-valued entries. These values must be encoded in a quantum state to train a quantum classifier.
    \item \textbf{Quantum data encoding phase}: In order to realize a BERT-Quant and an ELECTRA-Quant pipeline, embedding from pre-trained BERT and ELECTRA, need to be transformed in a quantum object, either a state vector or an operator to be used in the quantum computational pipeline. Several methods exist to encode classical data into quantum states: one possibility is to encode data into the complex amplitudes of a multi-qubit state. This approach requires that the $N$ dimensional feature vector is encoded in the superposition of $n=log_{2}N$ qubits. Additionally, the embeddings can be encoded as angular values of single qubit rotational gates. In this case, the $N$ dimensional feature vector requires $n=N$ qubits to be adequately encoded in a quantum state. The embeddings are generated by encoding the classical embeddings into quantum states, where the amplitudes represent the weights or probabilities associated with each word or document. Quantum circuits or operators perform transformations on the quantum state, enabling the manipulation and analysis of the quantum embeddings.
    \item \textbf{Quantum Classification Model}: Once the quantum embeddings are constructed, they can be used as inputs to quantum classification models for various NLP tasks. This work constructs the variational quantum classifier with Quantum-inspired neural networks (QNNs). These are generally composed of single and two-qubit gates with free parameters trained in the learning phase. The structure of these circuits needs to be sufficiently complex to accommodate the possible solutions of the task and yet simple enough to prevent detrimental effects from quantum noise. There is currently an ongoing research effort aiming to find the optimal strategy for the construction of variational quantum circuits \cite{Du2022}: one of the critical elements to take into account is a certain level of entanglement, i.e., quantum correlation, between qubits so that information is shared among each element of the computation and complex solution are potentially explored during training. Here, a basic entanglement ansatz is used, where each qubit will be forced into a quantum-correlated state with another qubit pairwisely. 
    \item \textbf{Model Training and Evaluation}: The quantum classification model and the quantum embeddings are trained using a labeled dataset. This dataset consists of instances with their corresponding class labels. Model training involves optimizing the parameters of the quantum classification model to minimize a predefined loss function. The trained model is evaluated using appropriate evaluation metrics, such as accuracy, precision, recall, and F1 score, to assess its performance on unseen data.
    \item \textbf{Evaluation}: To assess the effectiveness of quantum embeddings in QNLP tasks, comparing their performance against classical embedding-based models is beneficial. Classical models like traditional machine learning algorithms or neural networks can be trained and evaluated using the same dataset and evaluation metrics. The performance of the quantum embedding-based model can be compared to these classical models to determine the impact and benefits of quantum embeddings in QNLP tasks. It is worth noting that the specific implementation details of the methodology may vary depending on the chosen embedding technique, quantum embedding construction approach, and the specific quantum classification model employed. The above steps provide a general framework for integrating embeddings in Quantum Natural Language Processing.
\end{itemize}

\subsubsection{Quantum Amplitude Encoding.}
Classical data must be encoded in a quantum state to be manipulated in a quantum computational pipeline. One prevalent approach for this purpose is quantum amplitude encoding, representing classical data as amplitudes within a quantum superposition. This encoding method facilitates parallel computation and harnesses quantum interference for data processing tasks.
With quantum amplitude encoding, classical data are mapped onto quantum states by assigning complex amplitudes to specific computational basis states. More formally, consider here a classical dataset denoted by a feature vector $\mathbf{x} = (x_1, x_2, \ldots, x_N)$, where $N$ represents the dimensionality of the feature vector. The quantum amplitude encoding scheme transforms this feature vector into a quantum superposition state $|\mathbf{x}\rangle$.
To perform quantum amplitude encoding, a set of $n = \lceil \log_2 N \rceil$ qubits is utilized, with each qubit representing one element of the feature vector. The quantum state $|\mathbf{x}\rangle$ is given by:
\begin{equation}
|\mathbf{x}\rangle = \sum_{i=0}^{N-1} \alpha_i |i\rangle,
\end{equation}
where $|i\rangle$ denotes the $i$-th computational basis state of the $n$ qubits, and $\alpha_i$ represents the complex amplitude associated with each basis state. The amplitudes $\alpha_i$ are determined by the classical data values $x_i$.
Classical data are first normalized, i.e. the feature vector $\mathbf{x}$ is divided by the normalization factor $C$: this fact guarantees that the resulting quantum state is properly normalized, thus preserving the probabilistic interpretation of the wave function. Once the feature vector is normalized 
\begin{equation}
\mathbf{x}' = \frac{\mathbf{x}}{C},
\end{equation}
where $\mathbf{x}' = (x'_1, x'_2, \ldots, x'_N)$, the next step is the actual encoding phase. Here, the complex amplitudes $\alpha_i$ for each computational basis state $|i\rangle$ are given by:
\begin{equation}
\alpha_i = \frac{x'i}{\sqrt{\sum_{j=0}^{N-1} |x'_{j}|^2}},
\end{equation}
ensuring that the quantum state $|\mathbf{x}\rangle$ is properly normalized, with the sum of squared amplitudes equal to one.
The quantum amplitude encoding offers a means to transform classical data into quantum states, facilitating the utilization of quantum algorithms for data processing and machine learning tasks. 
\subsubsection{Parametrized Quantum Circuit.} The emebddings encoded in the quantum amplitudes of a statevector are processed by a parametrized quantum circut. A parametrized quantum circuit is technically a chain of quantum gates with free parameters \cite{benedetti2019parameterized}, which are iteratively updated to otpimize an objective function. In this sense, a parametrized quantum circuit in a learning optimization algorithm, is the quantum equivalent of a classical neural network. To exploit the computational and learning capabilities of parametrized quantum circuits it is necessary to put the qubits involved in an entangled state: these can be obtained with different approaches, and with varying topologies. There is no general procedure to infer which form of the entangling layers are more indicated for the specific task to be tackled: in this work a \textbf{Basic Entangling Layer} provided by Pennylane \cite{Plane}. The Basic Entangling Layer is composed of single parameter one qubit gate, applied on every qubit, and a ring of CNOT gates, i.e. two qubit gates connecting every qubit with the closest neighbour. The ring structure is given by the connection between the last qubit and the first one considered.

\subsection{Explainability with SHAP values}
To aid a qualitative analysis of the results discussed later in this work and extract a meaningful explanation about how a quantum transfer learning pipeline can be helpful in the Acceptability Judgement task, a SHAP value analysis is performed here. \textbf{SHAP} (SHapley Additive exPlanations) \cite{shapey} is a game-theoretical inspired approach to explain the output of an agnostic parametrized function, such as many deep learning models, using the inputs provided. In essence, SHAP values measure the relative contribution of each feature vector (in the task now considered, a single-word embedding) on the model's outcome. In this sense, SHAP values provide a quantitative mean for the local interpretability of an AI model. Hence, they are indirect quantifiers of the model's inner structure and computational reasoning. 
While the SHAP values are evaluated on single features, i.e., in the models considered here on single tokens, they can be clustered hierarchically in terms of the final score's co-occurring importance, revealing the overall contribution to the classification score of entire sections of sentences. This fact is of pivotal relevance for performing a qualitative analysis of the results of the models, as it allows to understand better which linguistic features impact correct classification and recurring patterns related to specific syntactic phenomena present in acceptable or unacceptable sentences. The methodology mentioned above for a qualitative analysis of the experimental findings is illustrated in more detail in the following section. 

\section{Results and Discussion}
\label{sec:results}
Two quantum transfer learning pipelines (BERT-Quant and ELECTRA-Quant) have been trained for the Acceptability Judgements on the Itacola Dataset. Similarly, for comparing the results, two classical transfer learning pipelines are considered (BERT-Classic and ELECTRA-Classic). For each of them, the main properties, hyperparameters and training strategies are listed in the following:

\begin{itemize}
    \item \textbf{BERT-Quant}: Emebdding extracted from BERT encoded in quantum state with the amplitude encoding strategy. To encode all the $768$ dimensions of the BERT feature vectors, the minimum number of qubit required is $n=\lceil log_{2}(768) \rceil= 10$, where the rest of the $1024$ amplitudes are padded all with $0.01$. The padding constant is non-informative, but its value needs to be selected in accordance with the chosen ansatz, as correlation phenomena between in-use qubit and not useful qubits might emerge during the computation. A rather simple but substantial empirical study has been conducted, revealing that a small but non zero value is more convenient for the learning process: this is due to the fact that some elements of the original feature vector need to be encoded in the portion of the Hilbert space spanned by the padding qubits \cite{Gianani}, hence a zero probability of occurrence would simply erase some relevant information too. Specifically, the value of the padding constant has been chosen to be on the same scale as the other amplitudes (normalized), to reduce at minimum the chances of biasing the result. After the encoding phase, the state obtained is used as an input for a $6$ layers of parametrized \textit{BasicEntangledLayer}. The outcome from the measurements, which consist in the expectation of Pauli-$Z$ for each qubit, are then passed to a Multi-Layer perceptron with a \textit{Softmax} activation function, and input-output dimension of $10-2$, trained for the classification, together with the entire circuit pipeline. The training is performed with an Adam optimizer, with leaning rate $10^{-5}$, batchsize of $32$ and a categorical cross entropy as objective function. The whole training process lasts for $7$ Epochs. 
    \item \textbf{ELECTRA-Quant}: Emebdding extracted from ELECTRA encoded in quantum state with the amplitude encoding strategy. To encode all the $768$ dimensions of the BERT feature vectors, the minimum number of qubit required is $n=\lceil log_{2}(768) \rceil= 10$, where the rest of the $1024$ amplitudes are padded all with $0.01$. After the encoding phase, the state obtained is used as an input for a $8$ layers of parametrized \textit{BasicEntangledLayer}. The outcome from the measurements, which consist in the expectation of Pauli-$Z$ for each qubit, are then passed to a Multi-Layer perceptron with a \textit{Softmax} activation function, and input-output dimension of $10-2$, trained for the classification, together with the entire circuit pipeline. The training is performed with an Adam optimizer, with leaning rate $10^{-5}$, batchsize of $32$ and a categorical cross entropy as objective function. The whole training process lasts for $7$ Epochs.
    \item \textbf{BERT-Classic} and \textbf{ELECTRA-Classic}: two classical transfer learning pipelines have been adopted for fine-tuning BERT and ELECTRA for sequence classification \cite{sun2019fine}. The training is performed with an Adam optimizer, with leaning rate $2\times 10^{-5}$ and eps $10^{-8}$, batchsize of $32$, $2$ as number of labels, $0$ as number of warm-up steps, $64$  as maximum length of the input sequences of words, and a categorical cross entropy as objective function. The whole training process lasts for $5$ Epochs. 
\end{itemize}
Concerning the quantum pipelines, the optimal number of layers for each model has been found empirically, taking into account the need for a trade-off between the performances and the computational cost \cite{Abbas2021}. While generally more layers are helpful in the learning process, they are computationally expensive. This goes together with the possibility of increasing the quantum noise in the computation as the number of gates increases. In the experiments here exposed, given the nature of the quantum hardware and the size of the problem, $6$ layers for the Basic Entangling Layer give the optimal classification performances.

For the sake of completeness, also the performance achieved by a baseline using LSTM with FasText embeddings \cite{trotta-etal-2021-monolingual-cross} has been reported. 

Evaluations were performed using two metrics: Accuracy, the most used measure used to evaluate acceptability on the GLUE benchmark and Matthews Correlation Coefficient (MCC) \cite{peters2018deep}, which is a correlation measure for Boolean variables mainly fitted when evaluating unbalanced binary classifiers.

\begin{figure}[h!]
\centering
\includegraphics[width=\textwidth]{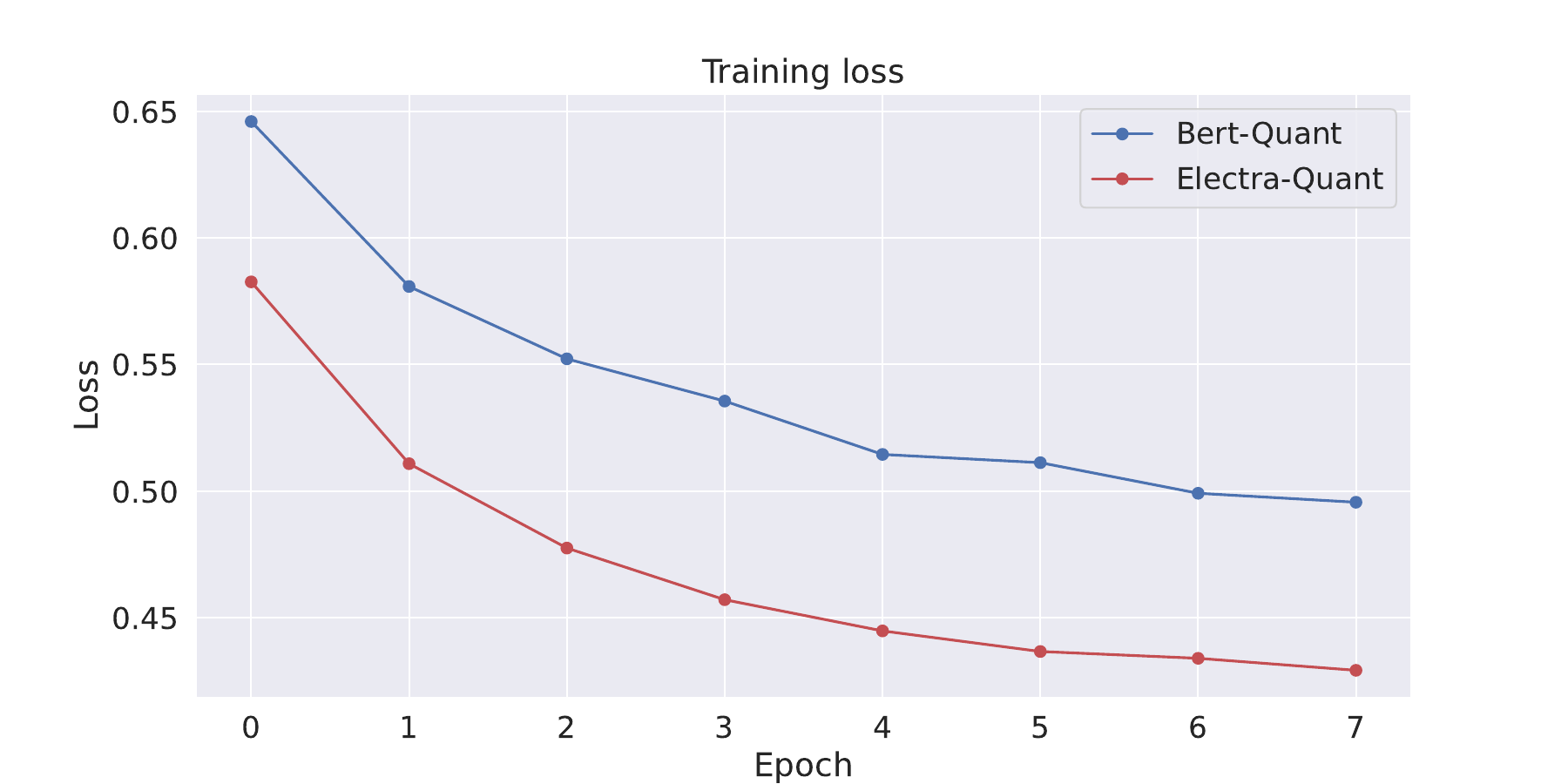}
\caption{\textbf{Loss function for the traing of BERT-Quant and ELECTRA-Quant.} The training phase of the two quantum transfer learning strategies constructed, namely BERT-Quant and ELECTRA-Quant, both with learning rate $10^-5$, batch size $16$ and categorical cross entropy as objective function. In both cases, the parametrized quantum circuit is made of $6$ Basic Entangling Layers. The training reveals that ELECTRA-Quant is a more effective for minimizing the loss, thus learns better than BERT-Quant to classify the acceptability of the sentences. }
\label{fig:loss}
\end{figure}

\subsection{Quantitative Analysis}

Concerning the quantum transfer learning pipelines, results of the various training are shown in Fig. \ref{fig:loss}. It is evident from the training phase that ELECTRA-Quant converges faster and to a lower objective function value than BERT-Quant, thus indicating the higher expressive value of the embedding extracted from the pre-trained ELECTRA model. This in fact, is confirmed by the accuracy and the MCC on the test set, as shown in table \ref{tab:baseline2}. Both BERT-Quant and ELECTRA-Quant outperform the LSTM model from the leaderboard, but BERT-Quant falls narrowly behind BERT-Classic in terms of accuracy and MCC. On the contrary, ELECTRA-Quant outperforms all the leaderboard models, with a mean accuracy of $0.92$ and a mean MCC of $0.676$. Concerning accuracy scores, these values are comparable with those achieved by ELECTRA-Classic using the classical approach ($0.923$), while the BERT-Classic stops at $0.904$.

\begin{table}[ht]
\centering
\begin{tabular}{|l|c|c|}

\hline
\textbf{Model} & \textbf{Accuracy} & \textbf{MCC} \\ \hline
LSTM & $0.794$ & $0.278 \pm 0.029$  \\ \hline
BERT-Classic & $0.904$ &$0.603 \pm 0.022$ \\ \hline
ELECTRA-Classic & $0.923 \pm 0.008$ & $0.690 \pm 0.035$ \\ \hline
\textbf{BERT-Quant} & $0.899 \pm 0.009$ & $0.575 \pm 0.072$\\ \hline
\textbf{ELECTRA-Quant} & $0.920 \pm 0.008 $ & $0.678 \pm 0.009$ \\ 
%\hline
 \hline
\end{tabular}
\caption{Comparison of classification results on the ItaCoLA test set using a classic approach based on LSTM, BERT and ELECTRA; and the quantum transfer learning pipelines: BERT-Quant and ELECTRA-Quant.} \label{tab:baseline2}
\end{table}

The result is even more significant considering the MCC as a metric, which is much better suited to this type of task \cite{peters2018deep}. In such case, although the best result is still achieved using ELECTRA-Classic with the classical approach, the deviation with ELECTRA-Quant is minimal (0.698 vs 0.676), confirming the potential of quantum circuits to represent language data in complex high dimensional vector spaces. This ability can fuel the most sophisticated NLP deep learning model with higher expressive capabilities, potentially impacting task performances beyond the binary classification hereby investigated.

\subsection{Qualitative Analysis}

Furthermore, a qualitative analysis of a sample extracted from the ItaCola dataset has also been conducted. In particular, we have selected a set of $100$ representative sentences showing different phenomena according to ItaCola expert annotations \cite{trotta-etal-2021-monolingual-cross}.

Listed below are the phenomena chosen to be included in the sample and why they were chosen:

\begin{itemize}

    \item \textbf{Simple}: Sentences that have a single verb and only the mandatory arguments, e.g., ``Il cane rosicchia un osso'' (En. \textit{The dog gnaws a bone.}) these are the most uncomplicated complexity sentences; they have a linear syntax and present no particular difficulty. The possible unacceptability of the sentence lies only in the violation of the order of the constituents.

    \item \textbf{Cleft constructions}: Sentences where a constituent has been moved to put it in focus, e.g. ``È Max che Maria ha invitato per cena'' (En. \textit{It is Max who Maria invited for dinner}.) This phenomenon is exciting because Italian language has an almost free word order \cite{bates1982functional}, unlike both English and other Romance languages such as French, which instead respect the canonical subject-verb-object (SVO) order. This great syntactic flexibility leads to complex syntactic constructions with inversion of constituents concerning the verbal head \cite{guarasci2020lexicon}.
    
    \item \textbf{Subject-verb agreement}: Sentences characterized by the presence or lack of subject and verb agreement in gender or number, e.g. ``Ho saputo che ieri la zia di Maria ti ha raccontato che Andrea ha incontrato Mauro'' (En. \textit{[I] heard that yesterday Maria's aunt told you that Andrea met with Mauro}.) Given the morphological richness of Italian and its relative freedom in constructing phrasal structures\cite{tsarfaty2010statistical}, verbal inflection determines the agreement between constituents and not their positional proximity. Various studies have correlated these inflectional properties with syntactic ones and phenomena such as the omission of the subject pronoun (pro drop)\cite{liu2012quantitative, bosco2012treebank}. Acceptability violations of sentences of this kind are both in the suffix indicating gender and number agreement and in the potential nested subordinate propositions.
\end{itemize}

To study how different phenomena impact differently and better understand how and what parts of a given syntactic structure impact classification, the visual formalism of the dendrogram was used. The choice is due to several reasons. First, the dendrogram is shap-compliant, allowing visual output to be easily interpreted and compared. Second, this work does not focus on an exclusively syntactic task, therefore, since no accurate syntactic analysis was performed on the sentences via the canonical dependency parse tree (DPT). Without a proper DPT - a mandatory step in every NLP pipelines - it is impossible to have a hierarchical structure labeled via the syntactic relations of each sentence. The dendrogram, therefore, is the best compromise, having already been used in some studies as an approximation of syntactic relations instead of DPT in tasks involving syntactic features \cite{sagae2009clustering}, in particular for rich-inflected languages \cite{bohnet2013joint}.

From the analysis performed on the sample under review, different behaviors are noted based on the phenomena that distinguish the sentences.
  
\begin{figure}[h]
\centering
\includegraphics[width=0.80\textwidth]{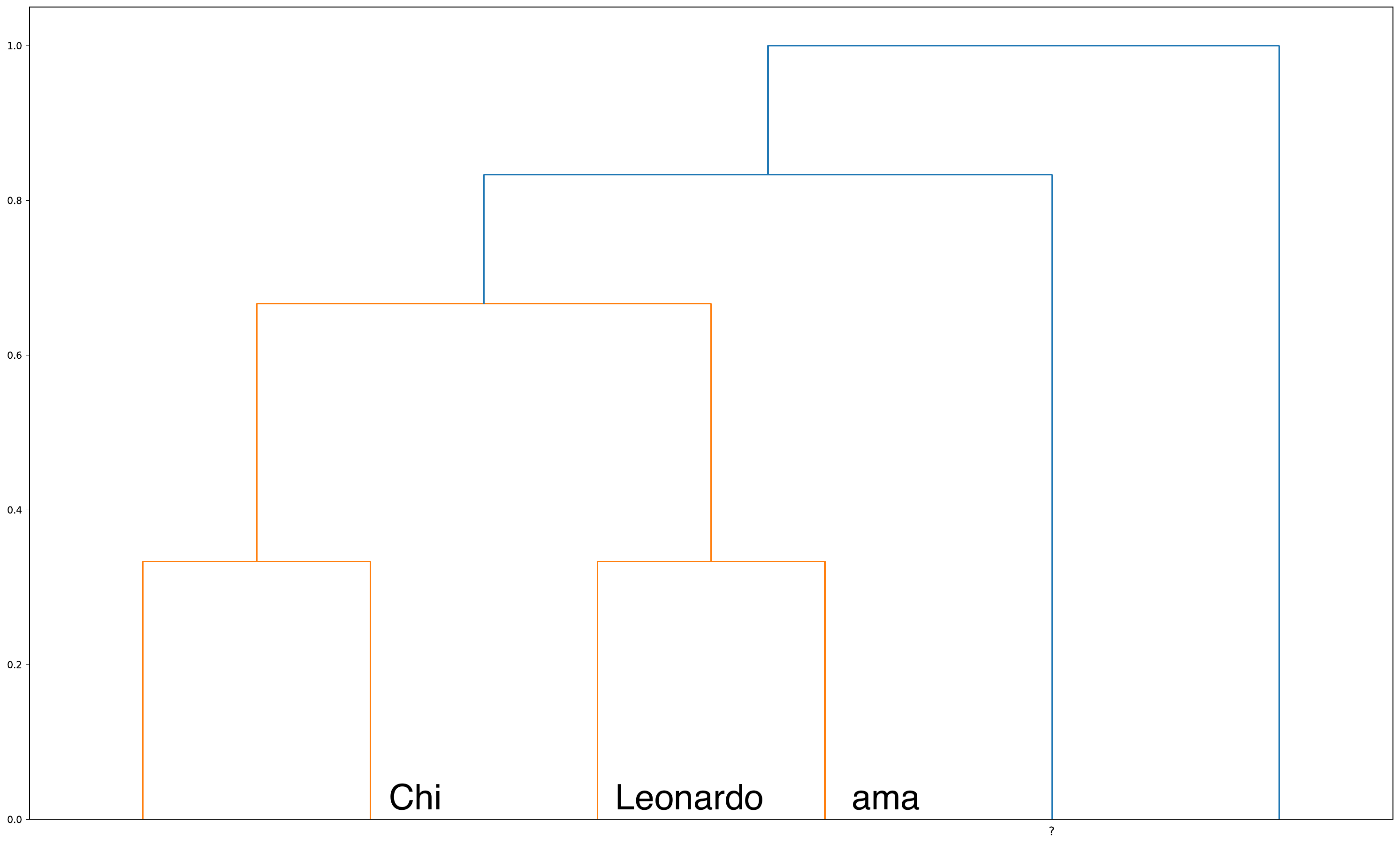}
\caption{Example of a dendrogram representation for a simple sentence in an interrogative form. Orange arcs represent the parts of the sentence that positively impact classification, classifying it as an unacceptable sentence. }
\label{fig:simpledendro}
\end{figure}

For instance, in the unacceptable sentence "Chi Leonardo ama?" (\textit{Who does Leonardo love?}) shown in figure \ref{fig:simpledendro}, arcs highlighted in orange represent the portions of the sentences (in this case the sentence as a whole) that contribute to the correct classification, i.e., classify the sentence as unacceptable. 

\begin{figure}[h]
\centering
\includegraphics[width=0.80\textwidth]{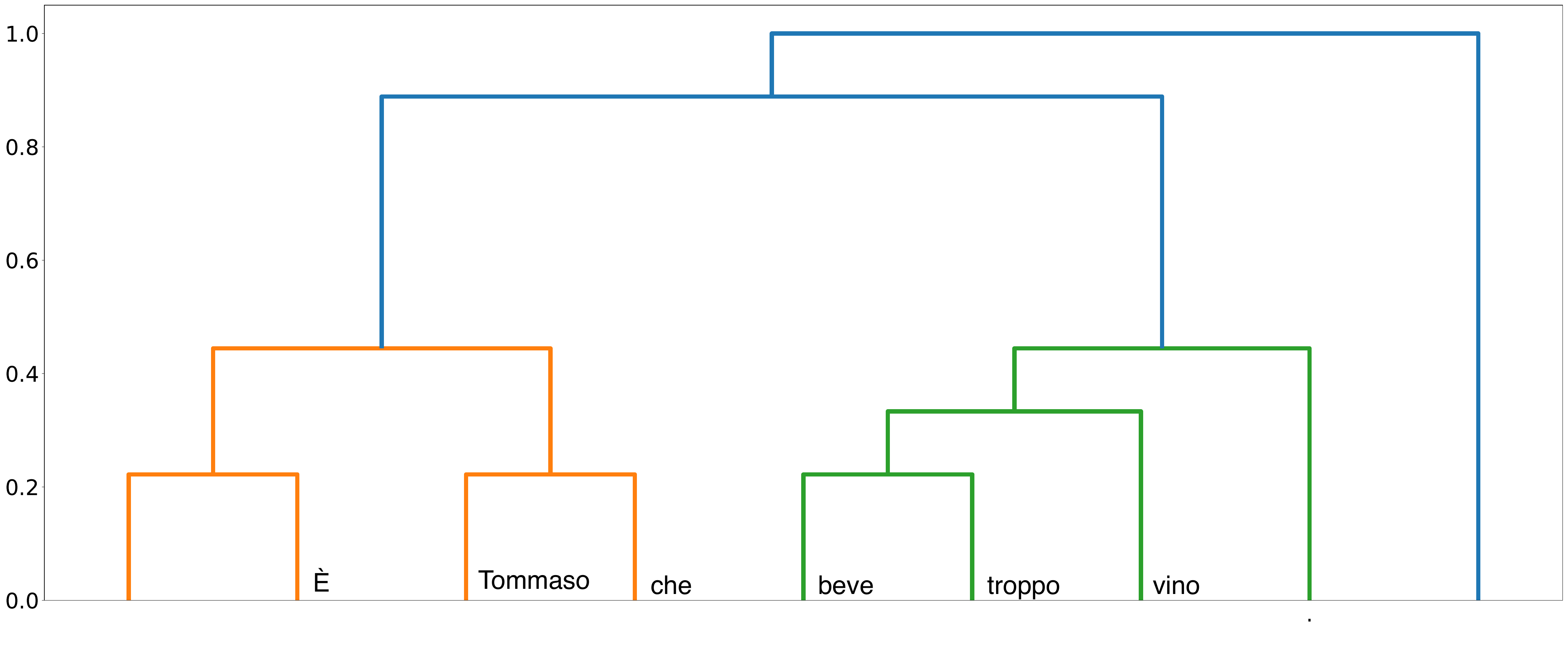}
\caption{Example of a acceptable sentence presenting a cleft structure. Orange arcs, representing correctly classified words, coincide with the main clause, which is also the most readable part, while the subordinate is the part that negatively impacts the correct classification of the sentence as acceptable (blue arcs). }
\label{fig:dendro_1_cleft}
\end{figure}

Moving to a more complex sentence, such as the one in Figure \ref{fig:dendro_1_cleft} "E' Tommaso che beve troppo vino" (\textit{It is Thomas who drinks too much wine}) arcs in two different colors can be seen. As mentioned above, green arcs indicate the phrase that contributes to classifying the sentence correctly, and the orange ones have a negative effect.
Analyzing this dendrogram in detail, it is possible to see that the portion of the sentence correctly influential for correct classification coincides \textit{de facto} with the main clause "beve troppo vino" (\textit{[He] drinks too much wine}). Notice that in this case, the main clause is placed after the dependent one "E' Tommaso che" (\textit{It is Thomas that}). This is because it is a cleft sentence whose peculiarity lies in having the order of constituents inverted, namely the dependent clause (usually a prepositional phrase) before the main clause.

The key to interpreting dendrograms can already be deduced from these two examples. In the first example, given a simple and unacceptable sentence, all the words that compose it impact the correct classification. In the second case, when dealing with an acceptable complex sentence, the portion that does not contribute to the correct identification is found in the less readable part of the sentence, namely, the prepositional dependent clause at the beginning. The criterion is, therefore, influenced not only by the judgment of acceptable/unacceptable but also by the readability of the sentence, expressed in terms of syntactic complexity.

\subsubsection{Quantum vs Classic Dendrograms}

From the analysis conducted on the sample, some differences in the behavior of classic and quantum models have been observed. In particular, the comparison has been focused on dendrograms produced by models that respectively achieved the best results, ELECTRA Classic and Quantum.
The behavior of the two different models concerning each of the phenomena included in the sample is briefly described below. 
%simple
As expected, both models find no difficulty handling simple sentences, showing similar classification behavior.

\begin{figure}[h]
\centering
\includegraphics[width=0.80\textwidth]{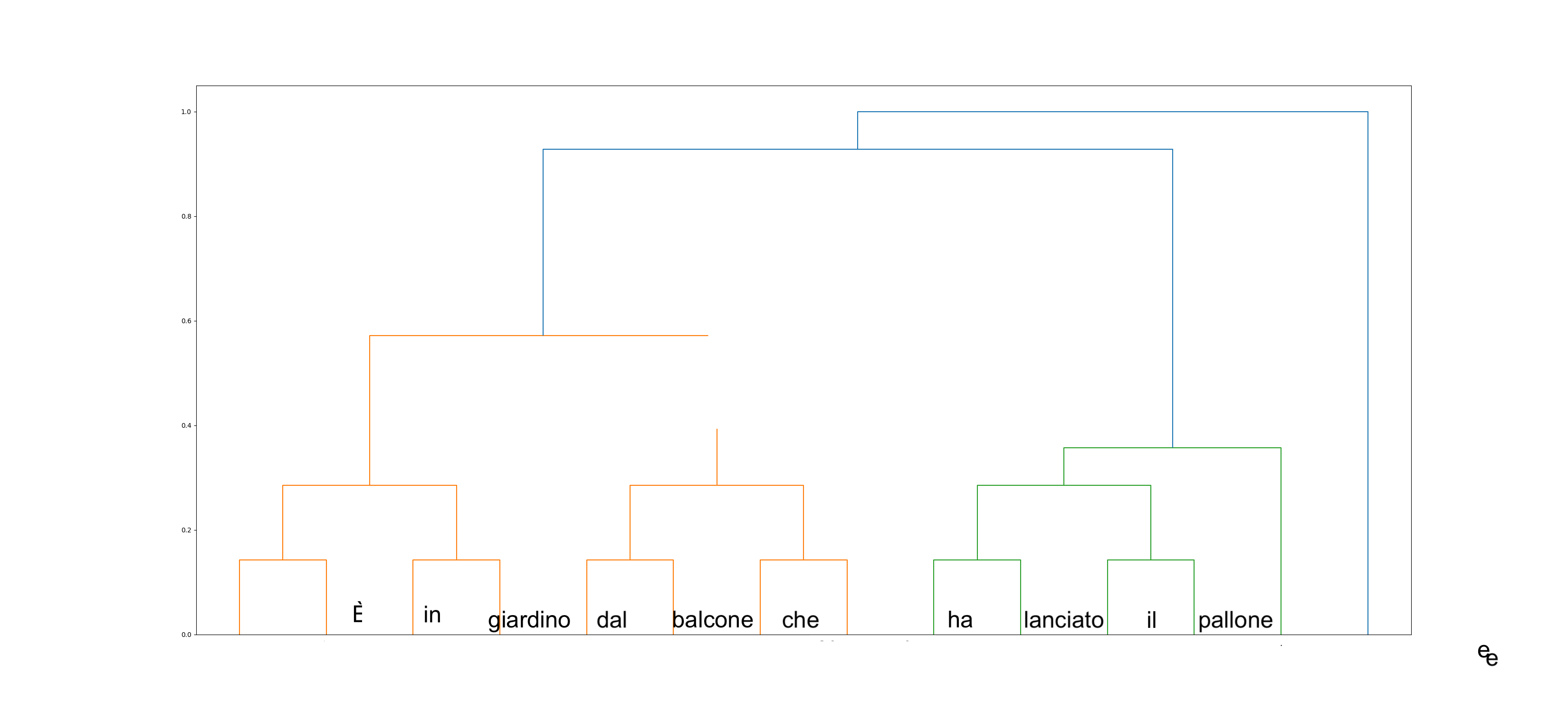}
\caption{Example of a sentence containing cleft construction phenomenon correctly classified. The main clause is identified by green arcs, while the relative clause placed before is in red.}
\label{fig:dendro_cleft_1}
\end{figure}

%cleft
The behavior of the two models about cleft constructions is more erratic.
For instance, specific moderately complex sentences such as the one depicted in Figure \ref{fig:dendro_cleft_1}, "\textit{è in giardino dal balcone che Alessandro ha lanciato il pallone}"(It is from the balcony that Alessandro has thrown the ball into the garden), are accurately classified only by Electra Classic. 

\begin{figure}[h]
\centering
\includegraphics[width=0.80\textwidth]{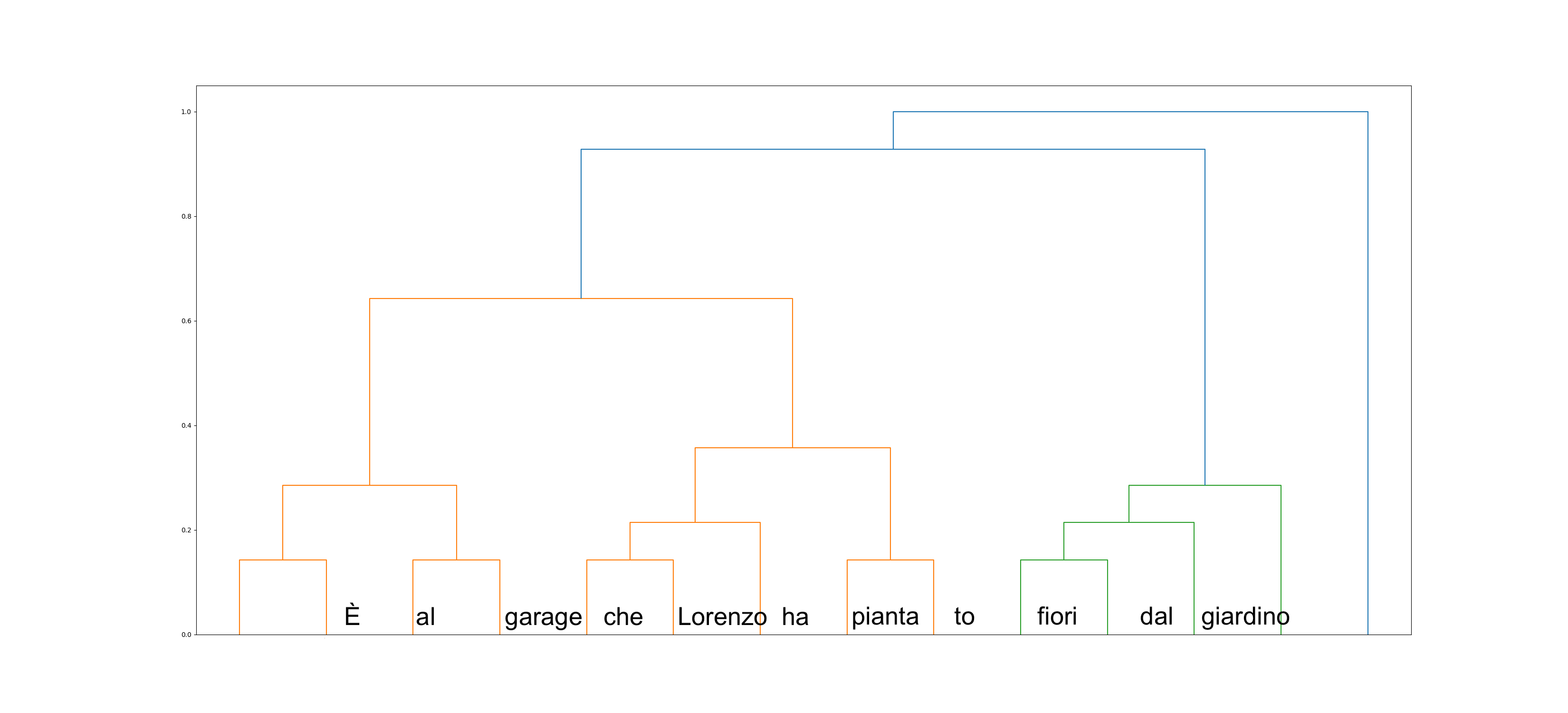}
\caption{Example of a cleft sentence wrongly classified. Constituents that impact classification are meaningless, and the verb is divided into two tokens.}
\label{fig:dendro_cleft_2}
\end{figure}

Conversely, the same model misclassifies straightforward sentences like the one shown in figure \ref{fig:dendro_cleft_2}:  \textit{"è al garage che Lorenzo ha piantato fiori dal giardino"} (It is in the garage that Lorenzo has planted flowers from the garden). In this case, the verb "\textit{piantare}" (to plant) is erroneously split, and the portion of the sentence that affects the classification is identified within a meaningless sequence lacking a verbal head, namely "\textit{fiori dal giardino}" (flowers from the garden).

Nevertheless, a recurring pattern does not appear to emerge: in some instances, the behavior is reversed, with Electra Quantum demonstrating improved performance on complex sentences and poorer performance on simpler ones, or vice versa.

%sva
In contrast, the situation is more precise in the case of sentences containing subject-verb agreements. This phenomenon, directly influenced by the rich inflectional morphology of Italian, can result in syntactically very complex sentences with various levels of embedded clauses in which the constituents' position makes readability difficult.

\begin{figure}[h]
\centering
\includegraphics[width=0.80\textwidth]{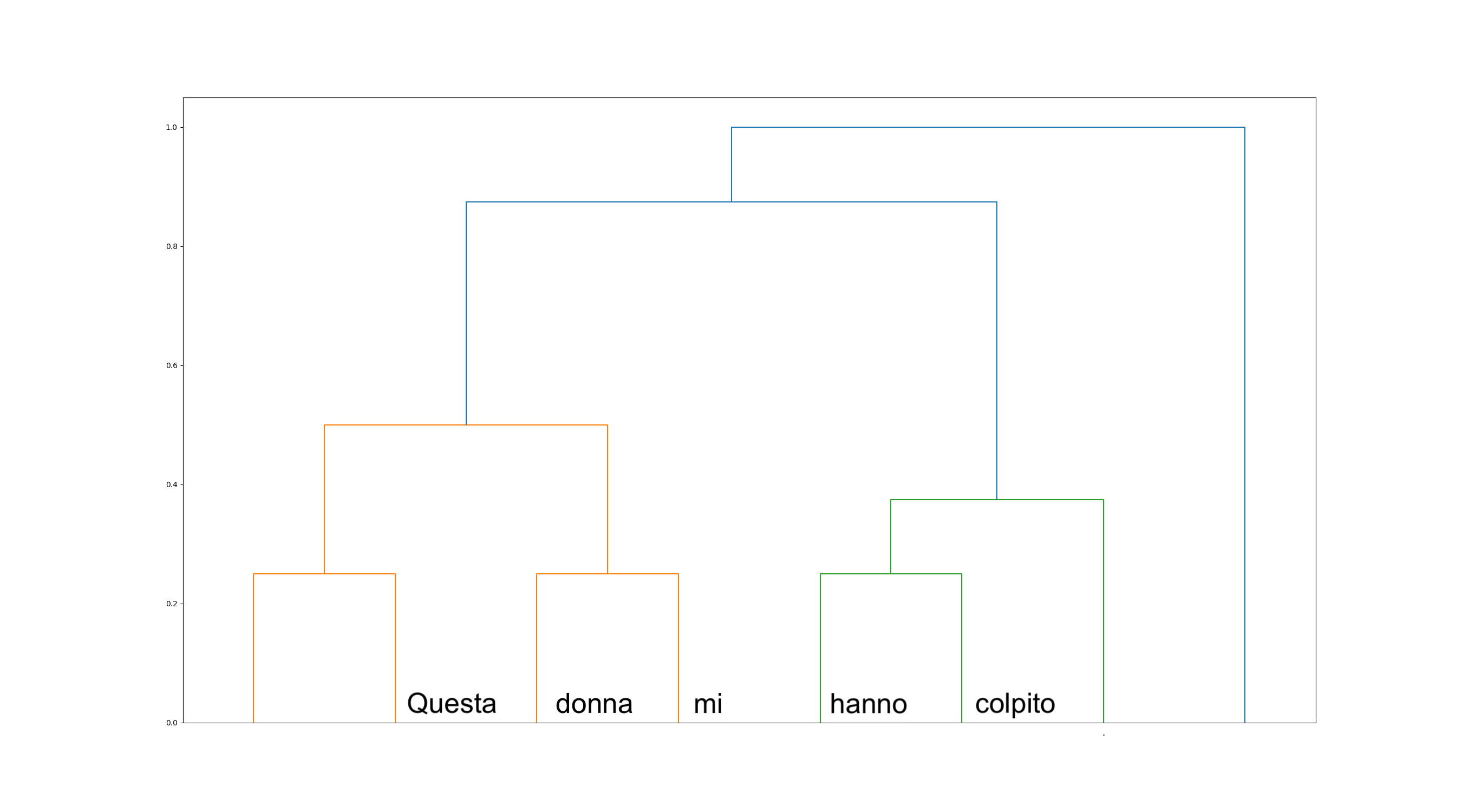}
\caption{Example of a sentence containing subject-verb-agreement phenomenon. }
\label{fig:dendro_sva}
\end{figure}

As shown in the figure \ref{fig:dendro_sva}, an unacceptable sentence such as \textit{"Questa donna mi hanno colpito"} (This woman they impressed me.) is handled identically by the models producing the same dendrograms, with the exact phrases concurring to impact positively or negatively in classifying.

\begin{figure}[h]
\centering
\includegraphics[width=0.80\textwidth]{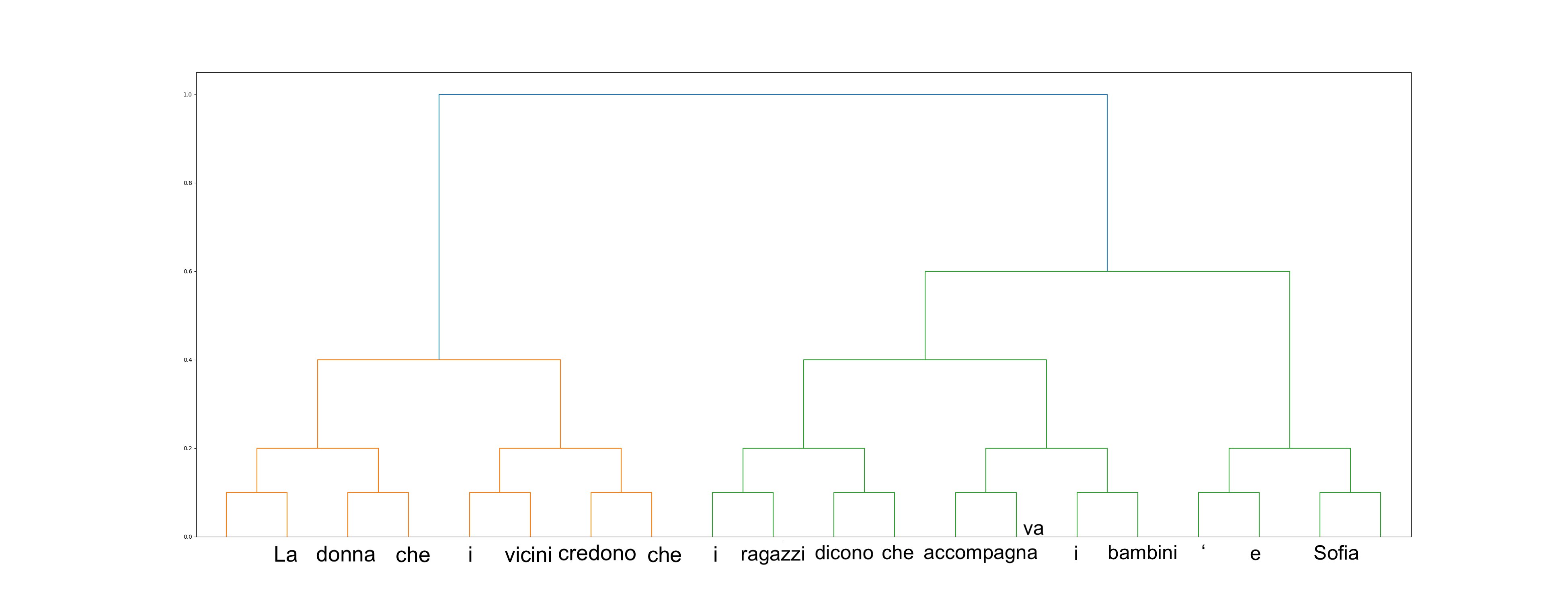}
22\caption{Example of a sentence containing subject-verb-agreement phenomenon with a high level of nested dependent clauses. }
\label{fig:dendro_sva2}
\end{figure}

However, by increasing the complexity of the sentence, the behavior changes. Figure \ref{fig:dendro_sva2} sentence is much less readable than the previous one: \textit{"La donna che i vicini credono che i ragazzi dicono che accompagnava i bambini è Sofia"} (The woman the neighbors believe the boys say was accompanying the children is Sofia). The sentence, while hostile even for a native speaker, is acceptable even though grammatically, it has a construction that pushes the dependent clauses nesting allowed by Italian to the maximum.

Electra quantum tends to make fewer errors and classify better in sentences such as this one, which is highly represented in the dataset. As can be seen from the figure, the green arcs that contribute to the correct interpretation approximate the independent clause. In contrast, the red arcs correspond to the less readable part consisting of a series of nested dependent clauses cascading between them.

While Electra quantum classify better a certain class of structured sentence, falls shortly behind the classical Electra model on simpler and less structured ones. This phenomenon is captured by the overall accuracy of the model: a net out-performance on any element of the dataset of the quantum model, would have led to higher accuracy or MCC in the classification, and this is clearly not the case. To understand the reason behind this, it is necessary to reason around the structure of the quantum circuit: the amplitude encoding in particular, here employed for writing Bert/Electra vectors onto a quantum state, while convenient in terms of qubits, is not entirely lossless as an encoding protocol: it has been demonstrated in fact that a growing number of queries is necessary to fully determine the belonging class of a feature vector encoded in the amplitude of a quantum state \cite{Wiebe_2020}. It follows that the single word embeddings loose partially their representation power. Hence, short sentences, where the weight of the single words representation is concurrent in importance with the syntactic structure for the acceptability, suffer more for the limitations posed by the encoding. The acceptability of long and structured sentences on the contrary, heavily depend on the correctness of the syntactic relationships, which are well captured by the parametrized ansatz, and thus they are correctly interpreted by the Electra quantum model. 

This is a fascinating behavior since Classical NLMs indeed have already shown excellent results in the literature in handling syntactic phenomena such as this one\cite{guarasci2021assessing,guarasci2022bert}. The improved performance offered by quantum models opens up new possibilities for more effective management of problematic phenomena pervasive in languages such as Italian and often challenging syntactic parsers due to convoluted syntax and sentence length \cite{brunato2018sentence}.

\section{Conclusion}
\label{sec:conclusion}

In this work, a novel approach for transfer learning with quantum computing has been investigated, focused on the Acceptability Judgments task on the Italian language using an annotated dataset. 

Specifically, using BERT and ELECTRA embeddings, two quantum classifier has been trained: BERT-Quant and ELECTRA-Quant. The training results demonstrate this approach's potential: the metrics of BERT-Quant outperform the LSTM model while falling shortly behind BERT-Classic. ELECTRA-Quant, on the contrary, is comparable to all the models mentioned above, both in terms of accuracy and of MCC. 
A qualitative linguistic analysis of the training results, aided by the SHAP dendrograms, has shown differences in behavior based on the syntactic phenomena characterizing the sentence. It suggests new effective quantum-based management methods for complex linguistic structures, which have always been challenging to manage in NLP in inflectional languages such as Italian.

It is a promising result not only because it envisions the possibility of quantum transfer learning outperforming all of the significant standard transfer learning approaches, even beyond the NLP domain. The most intriguing features revealed by the experiments and the analysis lie in the ability of quantum computers to map the syntactic structures and the functional connection between constituents, composing the whole sentence more efficiently and with higher expressive power. This behavior is probably due to the structure of Hilbert space and to the intrinsic tensor operation performed by a quantum computer: the isomorphism between language, specifically the Lambek grammar, and the quantum operation was already demonstrated by Bob \cite{coecke2010mathematical}. Future work will focus on the possibility of a more general isomorphism between syntactic properties and quantum computations beyond the specific task presented here and not limited to the Italian language.
%% acc
From a strictly linguistic point of view, the prediction and classification of acceptability judgments in NLP play a crucial role in improving the quality and fluency of various NLP applications, ranging from language models to language generation. By advancing the understanding and modeling of acceptability judgments, researchers aim to enhance the effectiveness and naturalness of human-computer interactions, making NLP systems more linguistically informed and aligned with human expectations.

%% qnlp
It is worth mentioning that while quantum computing shows promise for NLP classification tasks, the field is still in its nascent stages, and many challenges need to be addressed. These challenges include the limited availability of quantum hardware, the development of quantum algorithms that effectively exploit the characteristics of language data, and the need for scalable and efficient quantum-classical hybrid systems to process large-scale language datasets.
In summary, the exploration of quantum computing in NLP classification tasks offers a novel perspective on addressing language processing challenges. Quantum-inspired algorithms, quantum embeddings, and quantum machine learning models present exciting avenues for improving classification accuracy, semantic understanding, and computational efficiency. Further research and development in this interdisciplinary field have the potential to unlock new possibilities for quantum-assisted natural language processing.

\section{Acknowledgements}
\label{sec:ack}
We acknowledge financial support from the project PNR MUR project PE0000013-FAIR

% limite attuali approcci ibridi
%Currently, hybrid classical-quantum models are limited in applicability because the continuous switching between classical and quantum processing units dramatically affects training and inference speed.


 \bibliographystyle{unsrt}
 \bibliography{miabiblio}

\end{document}